# Exploit the potential of Multi-column architecture for Crowd Counting

Junhao Cheng[1], Zhuojun Chen[2], XinYu Zhang[1], Yizhou Li[3], Xiaoyuan Jing[1*]

[1]Wuhan University, Wuhan, China
[2]University of Dundee, Dundee, UK
[3]Tokyo Institute of Technology, Tokyo, Japan
[1]{ponmma, zhangxinyu}@whu.edu.cn, jingxy_2000@126.com, [2]georgechenzj@outlook.com, [3]yli@ok.sc.e.titech.ac.jp

## Abstract

*Crowd counting is an important yet challenging task in computer vision due to serious occlusions, complex background and large scale variations, etc. Multi-column architecture is widely adopted to overcome these challenges, yielding state-of-the-art performance in many public benchmarks. However, there still are two issues in such design: scale limitation and feature similarity. Further performance improvements are thus restricted. In this paper, we propose a novel crowd counting framework called Pyramid Scale Network (PSNet) to explicitly address these issues. Specifically, for scale limitation, we adopt three Pyramid Scale Modules (PSM) to efficiently capture multi-scale features, which integrate a message passing mechanism and an attention mechanism into multi-column architecture. Moreover, for feature similarity, a novel loss function named Multi-column variance loss is introduced to make the features learned by each column in PSM appropriately different from each other. To the best of our knowledge, PSNet is the first work to explicitly address scale limitation and feature similarity in multi-column design. Extensive experiments on five benchmark datasets demonstrate the effectiveness of the proposed innovations as well as the superior performance over the state-of-the-art. Our code is publicly available at:*

*https://github.com/oahunc/Pyramid_Scale_Network*

## 1. Introduction

In recent years, the field of crowd counting has grown at an astonishing pace. Part of the reason for such growth is the increasing public security need such as video surveillance and traffic control due to the rapid increase of population and the resulting urbanization. Another part is the rapid development of Convolutional Neural Network(CNN) based image recognition methods, which enable deep learning models to solve the intricate problem with high accuracy.

A robust crowd counting model is the one that generalizes to universal scenarios spanning through occlusion, complex background, scale variation and non-uniform distribution, etc(some samples are shown in Fig. 1).

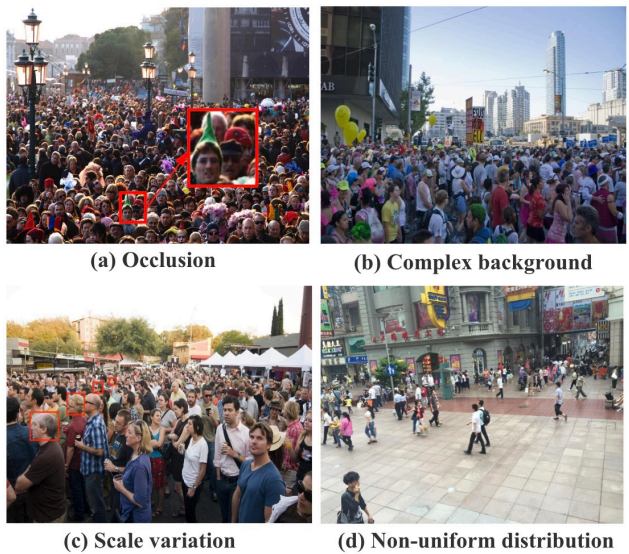

Figure 1. Representative images for challenges of occlusion, complex background, scale variation and non-uniform distribution.

Moreover, these challenges are not mutually exclusive. In other words, there may exist several challenges in one image. Among them, scale variation is the major challenge [34]. Hence, it is a continuing course to design more effective and robust frameworks to tackle these challenges.

Current mainstream methods take crowd counting as a regression task by outputting crowd density map whose sum is the number of people. Among them, some works have achieved significant successes by adopting multi-column architecture such as MCNN [4] and SANet [3]. The architecture usually divide the network into multiple columns to capture multi-scale features and combine their results at the output layer. However, these methods suffer two inherent algorithmic drawbacks: scale limitation and feature similarity. For the former, from a local perspective, each scale-specific column can only work well on its corresponding single scale and its performance will drop drastically on other scales, which is easy to result in poor results [35]. From a global perspective, the final total number of global scales corresponding to these methods equals the number of columns, which leads to the issue that these methods can only process crowd images with a few different scales. For the latter, it has been widely acknowledged that there is a lot of information redundancy in the multi-column architectures. The reason for information redundancy is that different columns seem to perform similarly without obvious differences, which have been demonstrated through conducting experiments by Li et al [2]. Based on this, we further hypothesize that different columns perform similarly due to feature similarity. Namely, in these multi-column methods, the output features derived by each column show only trivial differences between each other, which mitigates the merits of multi-column architecture. To prove our hypothesis, we use Cosine Similarity to measure the output feature similarity between different columns of MCNN [4] on ShanghaiTech Part_A dataset. Noting that Cosine Similarity ranges from [0,1], and the larger the value, the higher the similarity. It can be seen from Table 1 that there is a substantial feature similarity between each column, which indicates that each column learn nearly identical feature, thereby violating the original motivation of such multi-column architectures.

| Column | 1 VS 2 | 1 VS 3 | 2 VS 3 |
|---|---|---|---|
| Cosine Similarity | 0.8526 | 0.7932 | 0.9081 |

Table 1. Comparison of output feature similarity between columns in MCNN.

To deal with these two drawbacks, in this paper, we propose a Pyramid Scale Network (PSNet) for accurate crowd counting and high-quality density map generation, as shown in Fig. 2. Specifically, it solves these two drawbacks with pyramid scale representation learning and multi-column variance loss optimization. To go beyond scale limits, a Pyramid Scale Module (PSM) based on multi-column design is proposed to capture multi-scale features with a message passing mechanism and an attention mechanism. Namely, with a message passing mechanism, PSM can allow each column to correspond to multiple scales, which not only gives each column the capacity to extract multiple-scale features, but also increases the final total number of global scales. Further, inspired by the great success of attention models in many computer vision tasks [24], PSM uses an attention mechanism to weight the importance of the output multiple-scale feature of each column to select the most relevant piece of information for crowd counting. Noted that each column in the traditional multi-column architecture works independently, which further limits the performance improvement [36]. In contrast, PSM allows each column to work collaboratively, where different columns can communicate and exchange information with each other, which enables our model to implicitly mitigate feature similarity to some extent in our experiments. On the other hand, for ensuring feature variance of each column, we present a novel loss function named Multi-column variance loss to explicitly penalize similar signals in parallel columns corresponding to different scales in multi-column architectures, which is beneficial for further improving crowd counting estimation. To the best of our knowledge, PSNet is the first work to explicitly address scale limitation and feature similarity in multi-column design.

Extensive experimental results on five benchmark datasets demonstrate significant improvements of our method against the state-of-the-art methods on ShanghaiTech Part_A, Part_B [4], UCF_CC_50 [7] and UCF-QNRF [6] datasets and excellent performance on Mall [8] dataset. Ablation studies also demonstrate the effectiveness of each module within the proposed method.

To summarize contributions of this paper:
1) We propose a Pyramid Scale Module (PSM) for network design to solve the scale limitation issue.
2) We propose to use feature similarity measurement as loss function to improve feature diversity in multi-column architecture.
3) Extensive experimental results demonstrate that proposed PSNet outperforms the state-of-the-art crowd counting methods on several benchmarks.

## 2. Related works

Recent CNN-based crowd counting works can be categorized in to two groups: single- and multi-column. Single-column models has the architecture that is basically a deep convolution neural network, whilst the multi-column model is namely multiple columns corresponding to filters with different receptive field sizes applied to the input image to deal with different scaled humans, and the output features from these columns are fused to yield the final crowd counting estimation, which have brought about excellent performance for crowd counting. In the remaining part of this section, we briefly review single-column works, focusing on multi-column works.

**Single-column methods:** **Weighted-VLAD** [21] proposes to learn locality-aware feature (LAF) to extract both spatial and semantic information. **SaCNN** [23] is a scale-adaptive network which transforms the feature maps extracted from multiple layers to the same sizes and then combines them to generate the final density map. Besides, a global count loss is used to improve the network generalization on crowd scenes with few pedestrians. **CSRNet** [2] adopts dilated convolution layers to expand the receptive field and replace pooling operations. **SAA-Net** [25] is similar to SaCNN but it additionally uses a soft attention mechanism to fuse multi-scale features. **ADCrowdNet** [26] first uses an attention-aware network to detect crowd regions in images and compute the congestion degree of these regions, then employs a multi-scale deformable network to generate high-quality density maps. **DUBNet** [14] is a scalable and effective framework that can incorporate uncertainty quantification in prediction for crowd counting.

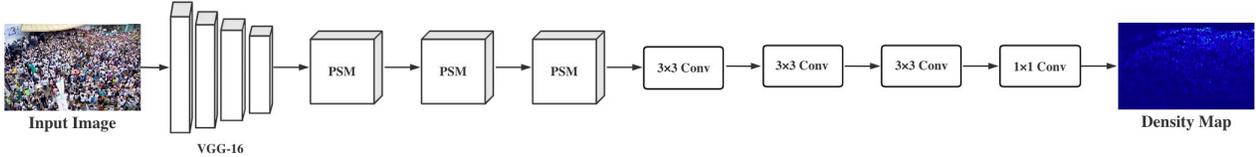

Figure 2. The architecture of the proposed Pyramid Scale Network(PSNet). The PSNet consists of a backbone network with the front ten layers of VGG-16, three stacked Pyramid Scale Modules(PSM), and four convolution layers for crowd density map regression. 3 × 3 Conv implies a convolution layer with 3 × 3 kernel size.

**Multi-column methods: MCNN** [4] is one of the earliest methods to adopt multi-column architecture to address scale variation in crowd counting, which comprises of three columns corresponding to filters with receptive fields of different sizes, designed for different human scales present in the images. **CrowdNet** [28] combines shallow and deep networks to predict the density map for a given crowd image, of which the shallow one captures the low-level features corresponding to large scale variation and the deep one captures the high-level semantic information. Similar to the above approaches, **Hydra-CNN** [27] consists of three heads and a body with each head learning features for a particular scale. Input to Hydra-CNN is pyramid of image patches corresponding to several scales which can directly expands the scale domain. **Switch-CNN** [10] inspired by MCNN employs an extra classification network to switch between multiple regressors for different scales, which can intelligently choose the most appropriate regressor for a particular input patch.

More recently, **SANet** [3] borrows Inception architecture [11] in the encoder to extract multi-scale features and uses a set of transposed convolution layers as the decoder to generate high-resolution density map. **TDF-CNN** [29] consists of a bottom-up network along with a separate top-down network. The bottom-up network, a two-column network, uses the feedback generated by the top-down network to correct the prior prediction in the retrospect. **DRSAN** [30] takes advantages of Spatial Transformer Network(STN) to handle scale variation and rotation variation. **SAAN** [31] utilizes an attention mechanism to automatically focus on certain global and local scales appropriate for the image. Similarly, **RANet** [32] introduces the idea of local and global self-attention which capture short-range and long-range interdependence information respectively. **DADNet** [33] takes dilated-CNN with various dilated rates to capture more contextual information, and utilizes adaptive deformable convolutions to generate a high-quality density map.

## 3. Method

### 3.1. PSNet Overview

Figure 2 shows the architecture of our proposed crowd counting framework called Pyramid Scale Network (PSNet). PSNet includes a backbone network as the feature extractor, three stacked Pyramid Scale Modules (PSM), and four convolution layers for final crowd density map regression.

The backbone network is based on VGG-16 [1] which is widely used for extracting low-level features such as texture and edges. Specifically, following previous works [2], we keep the front ten layers of VGG-16 with three pooling layers to be our backbone network considering the trade-off between resource cost and accuracy. Hence, the backbone network generates feature maps of 8 times smaller spatial width and height with respect to the input image. Noted that every subsequent process no longer changes the resolution of these feature maps.

The primary problem that should be addressed in crowd counting models is scale variation. As the scales of objects, i.e., the head sizes, vary according to the distance from the camera, multi-scale features are critical for distinguishing targets from background. To achieve this, PSNet adopts three stacked PSM to efficiently extract high-level multi-scale features. Then, these features are fed into four convolution layers for crowd density map regression.

Finally, we introduce a novel loss function named Multi-column variance loss for PSNet, which explicitly solve the feature similarity issue in multi-column architectures and further improve the learning capacity of the model.

### 3.2. Pyramid Scale Module

Pyramid Scale Module(PSM) adopts multi-column design to tackle the challenges of scale variation, which is partly motivated by the recent success of multi-column architectures [3, 33]. As discussed in Section 1, these state-of-the-art multi-column methods have the issue of scale limitation. Noted that the scale limitation here includes both local and global scale limitation. Namely, these methods can only extract features within a limited number of scales since each column corresponds to only one scale and the final total number of global scales corresponding to these methods equals the number of columns [35]. Moreover, independent work of each column further limits performance improvement [36].

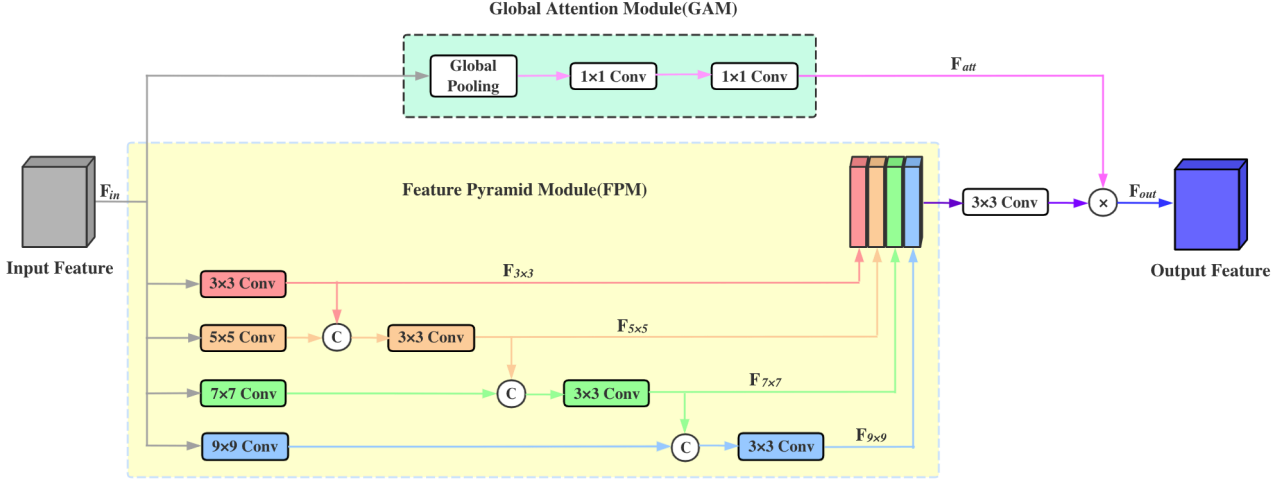

Figure 3. The architecture of the proposed Pyramid Scale Module(PSM). The PSM consists of Global Attention Module (GAM) and Feature Pyramid Module (FPM). GAM can provide global attention guidance, and FPM with four branches extracts multi-scale features through a message passing mechanism. C means concatenation, × denotes element-wise multiplication.

Based on the above observations, PSM applies a message passing mechanism and an attention mechanism to solve these issues. Specifically, PSM consists of two components: Global Attention Module (GAM) and Feature Pyramid Module (FPM), as shown in Fig. 3.

Like other attention models [5, 31], GAM provides global context attention as a guidance of multi-scale features generated by FPM to weight the importance of these features. Specifically, for the given input feature $F_{in} \in \mathbb{R}^{C \times H \times W}$ where $C$, $H$, $W$ respectively represent the channel, height and width of a input image, GAM performs global average pooling on $F_{in}$ and produce a channel vector $F_c \in \mathbb{R}^{C \times 1 \times 1}$ which softly encodes global context in each channel. To reduce parameter overhead, GAM first use a $1 \times 1$ convolution filter to reduce channels of $F_c$, and obtain $F_c \in \mathbb{R}^{C/r \times 1 \times 1}$, where r is the reduction ratio which is set to 16 in our experiments. Then GAM uses another $1 \times 1$ convolution filter to estimate attention across channels from $F_c$, and finally obtain global context attention $F_{att} \in \mathbb{R}^{C \times 1 \times 1}$.

FPM consists of four branches. For convenience of description, these four branches are called $3 \times 3$, $5 \times 5$, $7 \times 7$, $9 \times 9$ branch respectively. Following GAM, We also add a $1 \times 1$ convolution filter at the beginning of each branch to reduce the number of channels to 1/4 of its input feature. To build FPM with diverse scale, we use $3 \times 3$, $5 \times 5$, $7 \times 7$, $9 \times 9$ convolution filters for these four branches respectively. Besides, we can also consider the use of corresponding dilated convolution filters in substitution of these convolution filters, due to dilation convolution filter can performs similarly by enlarging receptive field without increasing the number of parameters and the number of computations [2]. Then, FPM integrates information of different scales step-by-step with a message passing mechanism. Namely, the concatenated output features from the lower branch and the adjacent upper branch are fused together using another $3 \times 3$ convolution filter as the final output feature of the lower branch. Next, a $3 \times 3$ convolution filter is used to fuse concatenated features from four branches. Finally, these multi-scale features are multiplied by the $F_{att}$ obtained in the GAM to get the final output feature $F_{out} \in \mathbb{R}^{C \times H \times W}$.

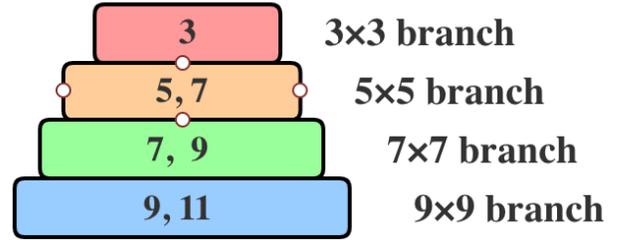

Figure 4. Illustration of scale diversity of corresponding branches of PSM.

Traditional multi-column methods have only one scale per column. On the contrary, from a local perspective, each branch of the PSM apart from the $3 \times 3$ branch corresponds to multiple scales, so these branches themselves have the capability to extract multi-scale features, as shown in Fig. 4. Moreover, from a global perspective, FPM also further enlarges the global scale range. Namely, the traditional four-column architecture corresponds to four scales, while FPM corresponds to five scales. In fact, each branch can use a convolution filter whose receptive field size is different from each other to fuse features, which will further increase the local and global scale diversity. The maximum numbers of scale for 3 × 3, 5 × 5, 7 × 7, and 9 × 9 branches are 1, 2, 3, 4 respectively. Moreover, FPM allows each branch to interact with each other, which implicitly overcomes the feature similarity issue to some extent and further enhances the representation learning capacity. In other words, FPM not only extracts more multi-scale features, but also allows each column to work in a collaborating way rather than a independent way.

### 3.3. Objective function

In Section 1, we experimentally proved in detail that the multi-column architectures suffer the feature similarity drawback. Specifically, the output features obtained by each column are very similar, which inevitably leads to a lot of information redundancy and further limits the capacity of these methods to extract multiple-scale features [2]. Indeed, this is the inherent drawback of multi-column architecture, which cannot be alleviated by simply modifying the multi-column architecture. Therefore, we propose a novel loss function named Multi-column variance loss to explicitly solve the feature similarity issue. Specifically, Multi-column variance loss forces each column to pay attention to different parts of the input feature as much as possible by making output feature of each column have obvious differences.

**Multi-column variance loss:** Namely, for the given output feature $F_{3\times3} \in \mathbb{R}^{C/4 \times H \times W}$ of 3 × 3 branch, we first apply average-pooling operation along the channel axis to obtain the corresponding attention map $F_{\text{map\_}3\times3} \in \mathbb{R}^{1 \times H \times W}$, then flatten $F_{\text{map\_}3\times3}$ to obtain the final attention vector $F_{\text{att\_}3\times3} \in \mathbb{R}^M$. The above series of operations can be expressed as:

$$F_{3\times3} \in \mathbb{R}^{C/4 \times H \times W} \rightarrow F_{\text{att\_}3\times3} \in \mathbb{R}^M \quad (1)$$

where M is $H \times W$. Similarly, attention vectors $F_{\text{att\_}5\times5}$, $F_{\text{att\_}7\times7}$ and $F_{\text{att\_}9\times9}$ of other three branches can be obtained.

Then, we adopt Cosine Similarity to measure the similarity of these four attention vectors. Therefore, we can define our Multi-column variance loss as:

$$L_M = \frac{1}{N \cdot K \cdot S} \sum_{n=1}^{N} \sum_{k=1}^{K} \sum_{s=1}^{S} \frac{F_{\text{att\_}ks} \cdot [(F_{\text{att\_sum}} - F_{\text{att\_}ks})/(S-1)]}{\max(|F_{\text{att\_}ks}| \cdot |(F_{\text{att\_sum}} - F_{\text{att\_}ks})/(S-1)|, \epsilon)} \quad (2)$$

where N is the number of image in the batch, K is the number of PSM in the PSNet, S is the number of branches in a PSM, $F_{\text{att\_}ks}$ is the attention vector of the corresponding branch. $|F_{\text{att\_}ks}|$ represents the norm of the attention vector. $\epsilon$ is a small value to avoid division by zero, which is set to $1e^{-6}$. $F_{\text{att\_sum}}$ is the sum of attention vectors corresponding to all branches in a PSM. The formula is as follow:

$$F_{\text{att\_sum}} = \sum_{s=1}^{S} F_{\text{att\_}ks} \quad (3)$$

Noting that the value range of $L_M$ is [0, 1], and the larger the $L_M$, the higher the similarity.

**Euclidean loss:** We choose the Euclidean distance to measure the estimation difference at pixel level between the estimated density map and the ground truth density map. The loss function is defined as follow:

$$L_E = \frac{1}{N} \sum_{n=1}^{N} |G(X_i; \theta) - D_i^{GT}|_2^2 \quad (4)$$

where $\theta$ indicates the model parameters, $X_i$ denotes the input image, $G(X_i; \theta)$ is the generated density map and $D_i^{GT}$ is the ground truth density map.

**Final objective:** By weighting the above two loss functions, the final objective function for training can be formulated as:

$$L = L_E + \lambda L_M \quad (5)$$

where $\lambda$ is the weight to balance the Multi-column variance loss and Euclidean loss. According to our experiments, its setted values on five datasets are shown in Table 2.

| Dataset | Value of weight |
|---------|-----------------|
| ShanghaiTech Part_A [4] | 1 |
| ShanghaiTech Part_B [4] | 0.01 |
| UCF_CC_50 [7] | 0.001 |
| UCF-QNRF [6] | 1.5 |
| Mall [8] | 0.001 |

Table 2. The values of weight $\lambda$ for different datasets.

### 3.4. Implementation details

Similar to other works [4], we generate the ground truth by way of blurring each head annotation with a Gaussian kernel(which is normalized to 1). Furthermore, for the crowded datasets including ShanghaiTech Part_A [4], UCF-QNRF [6] and UCF_CC_50 [7], we adopt the geometry-adaptive kernels to generate the ground truth, while the fixed Gaussian kernels are used to generate density maps for the datasets with relatively sparse crowd including ShanghaiTech Part_B [4] and Mall [8].

First ten layers of VGG-16 [1] model pretrained on ImageNet [37] is used to initialise the backbone network. All new layers are initialized from a Gaussian distribution with zero mean and 0.01 standard deviation. We train our PSNet in an end-to-end manner from scratch, and optimize the network parameters with Adam [9] optimizer with an initial learning rate of 1e-4.

Furthermore, during training, we firstly randomly scale the image with ratio [0.8, 1.2]. Then, images patches with fixed size are cropped at random locations. Coming processes include random mirroring with probability 0.5 and gamma contrast transform using parameter [0.5, 1.5] with probability 0.3. Also, we randomly change the color images to gray with probability 0.1 for the ShanghaiTech Part A [4] and UCF-QNRF [6] datasets that contain gray images. At test time, we neither scale nor crop the image and instead we feed whole image to the network.

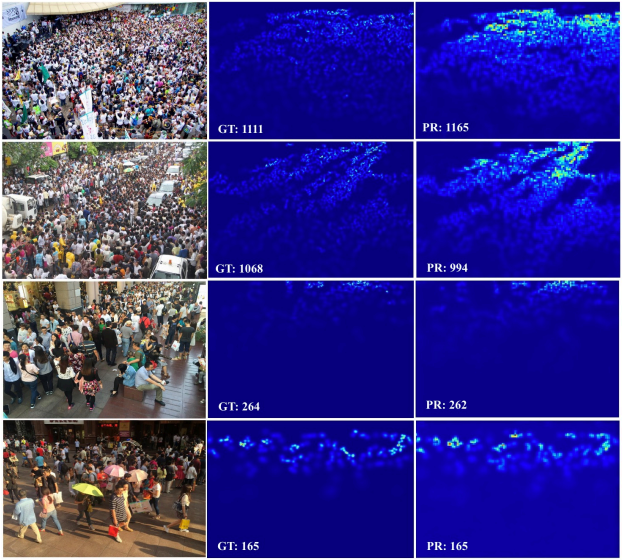

Figure 5. Visualization results of PSNet on ShanghaiTech dataset. For each group of images, pictures in the middle and on the right are corresponding ground truth and estimated density map of the image on the left.

## 4. Experiments

### 4.1. Evaluation metrics

We use mean absolute error (MAE) and root mean squared error (RMSE) for evaluating the network performance. These two metrics are defined as follows:

$$MAE = \frac{1}{N}\sum_{i=1}^{N}|C_i - C_i^{GT}| \qquad (6)$$

$$RMSE = \sqrt{\frac{1}{N}\sum_{i=1}^{N}|C_i - C_i^{GT}|^2} \qquad (7)$$

where $N$ is the total number of test images, $C_i$ and $C_i^{GT}$ represent the predicted count and ground truth count respectively. Roughly speaking, MAE determines the accuracy of the estimates, while RMSE indicates the robustness of the estimates.

### 4.2. ShanghaiTech

ShanghaiTech dataset [4] is composed of 1198 images with 330,165 annotations. According to different density distributions, the dataset is divided into two parts: Part_A with 482 images and Part_B with 716 images. Images in Part_A are more crowded than in Part_B. Both parts are further divided into train and test sets with train set of Part_A containing 300 images and that of Part_B containing 400 images. As shown in Table 3, our model achieves state-of-the-art performance on both datasets, where MAE is improved by 10.9% for Part_A. In addition, Fig. 5 presents some visual results of our approach on Part_A and Part_B dataset. The predicted density map exhibits good clarity and accuracy in both crowded and sparse areas.

| Method | Part A MAE | Part A RMSE | Part B MAE | Part B RMSE |
|---|---|---|---|---|
| MCNN [4] | 110.2 | 173.2 | 26.4 | 41.3 |
| Switching CNN[10] | 90.4 | 135.0 | 21.6 | 33.4 |
| CP-CNN [11] | 73.6 | 106.4 | 20.1 | 30.1 |
| ic-CNN [12] | 68.5 | 116.2 | 10.7 | 16.0 |
| CSRNet [2] | 68.2 | 115.0 | 10.6 | 16.0 |
| SANet [3] | 67.0 | 104.5 | 8.4 | 13.6 |
| DUBNet [14] | 64.6 | 106.8 | 7.7 | 12.5 |
| CAN [13] | 62.3 | 100.0 | 7.8 | 12.2 |
| PSNet (ours) | **55.5** | **90.1** | **6.8** | **10.7** |

Table 3. Performance comparison on ShanghaiTech dataset.

### 4.3. UCF_CC_50

The UCF_CC_50 dataset [7] contains 50 images of various resolutions and perspectives crawled from the internet. The number of annotated persons per image ranges from 94 to 4543 with an average number of 1280. We follow the standard setting in [7] to perform 5-fold cross-validation. As shown in Table 4, we achieve an improvement of 12.7% in terms of the MAE metric.

| Method | MAE | RMSE |
|---|---|---|
| MCNN [4] | 377.6 | 509.1 |
| Switching CNN [10] | 318.1 | 439.2 |
| CP-CNN [11] | 295.8 | 320.9 |
| ic-CNN [12] | 260.9 | 365.5 |
| CSRNet [2] | 266.1 | 397.5 |
| SANet[3] | 258.4 | 334.9 |
| DUBNet[14] | 243.8 | 329.3 |
| CAN[13] | 212.2 | **243.7** |
| PSNet (ours) | **185.3** | 265.0 |

Table 4. Performance comparison on UCF_CC_50 dataset.

### 4.4. UCF-QNRF

The UCF-QNRF dataset, introduced recently by Idrees et al [6], is a large and high-resolution dataset, which consists of 1,535 challenging images with 1.25 million annotations. The train and test sets in this dataset consist of 1201 and 334 images respectively. The UCF-QNRF dataset contains buildings, vegetation, sky and roads as they are present in realistic scenarios captured in the wild, which makes this dataset more realistic as well as difficult. Although it is considered a harder one, it can be seen from Table 5 that our method gains relative MAE improvement of 15.6%.

| Method | MAE | RMSE |
|---|---|---|
| Idrees et al. [15] | 315 | 508 |
| MCNN [4] | 277 | 426 |
| Switching CNN [10] | 228 | 445 |
| Resnet101 [16] | 190 | 277 |
| Densenet201 [17] | 163 | 226 |
| CL [6] | 132 | 191 |
| CAN [13] | 107 | 183 |
| DUBNet [14] | 105.6 | 180.5 |
| PSNet (ours) | **89.1** | **151.3** |

Table 5. Performance comparison on UCF-QNRF dataset.

### 4.5. Mall

Mall [8] dataset has 2000 annotated frames of over 60000 pedestrians captured from a shopping mall. We use the first 800 frames for training and the remaining 1200 frames for evaluation. Mall is not as challenging as others due to crowds in the dataset are sparse and all the images are the same scene. Compared with previous methods, our method achieves excellent performance with respect to both MAE and MSE. The results are shown in Table 6, which indicates that our method is also robust to sparse and invariant scene.

| Method | MAE | RMSE |
|---|---|---|
| R-FCN [18] | 6.02 | 5.46 |
| Faster R-CNN [19] | 5.91 | 6.60 |
| COUNT Forest [20] | 4.40 | 2.40 |
| Weighted VLAD [21] | 2.41 | 9.12 |
| DecideNet [22] | **1.52** | **1.90** |
| PSNet (ours) | 1.53 | 1.94 |

Table 6. Performance comparison on Mall dataset.

### 4.6. Ablation Study

**Effectiveness of various components and Multi-column variance loss:** We build multiple variants of PSNet by adding these components incrementally. The first variant is the **Baseline** trained only with Euclidean loss, which consists of a backbone network, three stacked FPM without the message passing mechanism, and four convolution layers. Furthermore, **Baseline** can be regarded as a traditional multi-column architecture. Then we adopt FPM with the message passing mechanism to build the second variant, **Baseline-FPM**. Next, we introduce the attention mechanism into the FPM, that is, replace the FPM with PSM, so as to obtain the third variant, **Baseline-PSM**. Finally, using Euclidean loss and Multi-column variance loss to jointly train the third variant, we obtained our final network, **PSNet**.

It can be seen from Table 7 that MAE, MSE and $L_M$ gradually decrease, that is, the model performance gradually improves, which indicates each part in our approach is effective and complementary to each other.

Moreover, the results evidence that message passing mechanism and Multi-column variance loss have the greatest impact on performance. Besides, it is worth noting that $L_M$ of the **Baseline** is very large, which further proves the severe information redundancy in the traditional multi-column design, in other words, the feature similarity mentioned in this paper. Moreover, compared with **Baseline**, the $L_M$ of **Baseline-FPM** has slightly improved, which validates our hypothesis that the message passing mechanism not only enhance the capacity of such multi-column architecture to extract multi-scale features, but also implicitly suppresses the feature similarity to a certain extent.

| Method | MAE | RMSE | $L_M$ |
|---|---|---|---|
| Baseline | 60.7 | 100.6 | 0.9612 |
| Baseline-FPM | 57.1 | 96.6 | 0.9425 |
| Baseline-PSM | 56.8 | 94.9 | 0.9367 |
| PSNet | 55.5 | 90.1 | 0.6116 |

Table 7. Performance comparison of different variants of PSNet on ShanghaiTech Part_A dataset.

**Scale Invariance:** We divide the ShanghaiTech Part_A test set into 10 groups on the basis of the crowd count. Each group can represent a specific scale level due to the dataset has large scale variations. Noted that we average the crowd count across a group to obtain the average count. The experimental results in Figure 6 show that the difference between the average counts for PSNet and GT is small for most groups, which further demonstrates the scale generalization of our mehod.

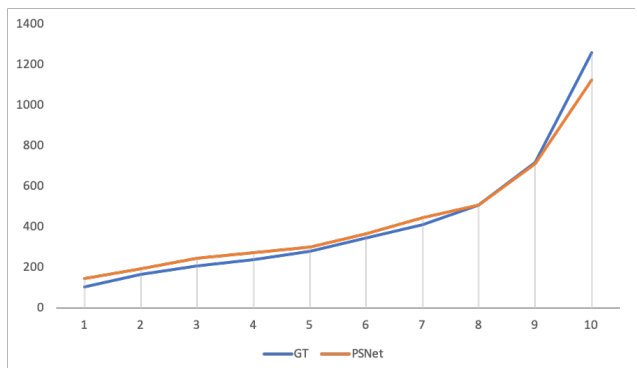

Figure 6. Performance across different scale level. GT denotes the ground truth, PSNet represents prediction of our method.

**Generalization of Multi-column variance loss**: We compare the experimental results of MCNN [4] with and without Multi-column variance loss training. Compared results are showed in Table 8. We observe that adopting Multi-column variance loss results in MAE and MSE are improved by 5.5% and 5.0% respectively, which shows that Multi-column variance loss can be generalized well to other existing multi-column architectures and can direcly applied to improve performance without any other changes. More remarkably, designing more efficient Multi-column variance loss is our future research work.

| Method | MAE | RMSE | $L_M$ |
|---|---|---|---|
| MCNN($L_E$) | 113.6 | 178.1 | 0.8836 |
| MCNN($L_E$+ $L_M$) | 107.3 | 169.2 | 0.7970 |

Table 8. Performance comparison of MCNN on ShanghaiTech Part_A dataset.

**Impact of dilation convolution:** As discussed in 3.2, we can also use the dilation convolution to replace the standard convolution in the FPM to further reduce parameter overhead. However, there is a slight drop in the result, but it still exceed the state-of-the-art methods, see Table 9.

| Method | MAE | RMSE |
|---|---|---|
| PSNet-dilation | 56.6 | 92.5 |
| PSNet | 55.5 | 90.1 |

Table 9. Performance comparison of PSNet with and without dilation convolution on ShanghaiTech Part_A dataset.

## 5. Conclusion

In this paper, we first discuss the two major drawbacks of the current multi-column methods: scale limitation and feature similarity. Then, we propose to use Multi-column variance loss and PSM with a message passing mechanism and an attention mechanism to overcome the above two limitations. Experiment results evidence the effectiveness of our method. More remarkably, our method can be easily applied to other existing multi-column networks. Our code is publicly available at:

https://github.com/oahunc/Pyramid_Scale_Network

## References

[1] Simonyan, Karen, and Andrew Zisserman. "Very deep convolutional networks for large-scale image recognition." arXiv preprint arXiv:1409.1556 (2014).
[2] Li, Yuhong, Xiaofan Zhang, and Deming Chen. "Csrnet: Dilated convolutional neural networks for understanding the highly congested scenes." Proceedings of the IEEE conference on computer vision and pattern recognition. 2018.
[3] Cao, Xinkun, et al. "Scale aggregation network for accurate and efficient crowd counting." Proceedings of the European Conference on Computer Vision (ECCV). 2018.
[4] Zhang, Yingying, et al. "Single-image crowd counting via multi-column convolutional neural network." Proceedings of the IEEE conference on computer vision and pattern recognition. 2016.
[5] Hu, Jie, Li Shen, and Gang Sun. "Squeeze-and-excitation networks." Proceedings of the IEEE conference on computer vision and pattern recognition. 2018.
[6] Idrees, Haroon, et al. "Composition loss for counting, density map estimation and localization in dense crowds." Proceedings of the European Conference on Computer Vision (ECCV). 2018.
[7] Idrees, Haroon, et al. "Multi-source multi-scale counting in extremely dense crowd images." Proceedings of the IEEE conference on computer vision and pattern recognition. 2013.
[8] Chen, Ke, et al. "Feature mining for localised crowd counting." BMVC. Vol. 1. No. 2. 2012.
[9] Kingma, Diederik P., and Jimmy Ba. "Adam: A method for stochastic optimization." arXiv preprint arXiv:1412.6980 (2014).
[10] Sam, Deepak Babu, Shiv Surya, and R. Venkatesh Babu. "Switching convolutional neural network for crowd counting." 2017 IEEE Conference on Computer Vision and Pattern Recognition (CVPR). IEEE, 2017.
[11] Szegedy, Christian, et al. "Going deeper with convolutions." Proceedings of the IEEE conference on computer vision and pattern recognition. 2015.
[12] Ranjan, Viresh, Hieu Le, and Minh Hoai. "Iterative crowd counting." Proceedings of the European Conference on Computer Vision (ECCV). 2018.
[13] Liu, Weizhe, Mathieu Salzmann, and Pascal Fua. "Context-aware crowd counting." Proceedings of the IEEE Conference on Computer Vision and Pattern Recognition. 2019.
[14] Oh, Min-hwan, Peder A. Olsen, and Karthikeyan Natesan Ramamurthy. "Crowd Counting with Decomposed Uncertainty." AAAI. 2020.
[15] Idrees, Haroon, et al. "Multi-source multi-scale counting in extremely dense crowd images." Proceedings of the IEEE conference on computer vision and pattern recognition. 2013.
[16] He, Kaiming, et al. "Deep residual learning for image recognition." Proceedings of the IEEE conference on computer vision and pattern recognition. 2016.
[17] Huang, Gao, et al. "Densely connected convolutional networks." Proceedings of the IEEE conference on computer vision and pattern recognition. 2017.
[18] Dai, Jifeng, et al. "R-fcn: Object detection via region-based fully convolutional networks." Advances in neural information processing systems. 2016.
[19] Ren, Shaoqing, et al. "Faster r-cnn: Towards real-time object detection with region proposal networks." Advances in neural information processing systems. 2015.
[20] Pham, Viet-Quoc, et al. "Count forest: Co-voting uncertain number of targets using random forest for crowd density estimation." Proceedings of the IEEE International Conference on Computer Vision. 2015.
[21] Sheng, Biyun, et al. "Crowd counting via weighted VLAD on a dense attribute feature map." IEEE Transactions on Circuits and Systems for Video Technology 28.8 (2016): 1788-1797.
[22] Liu, Jiang, et al. "Decidenet: Counting varying density crowds through attention guided detection and density estimation." Proceedings of the IEEE Conference on Computer Vision and Pattern Recognition. 2018.
[23] L. Zhang, M. Shi, and Q. Chen, "Crowd counting via scale-adaptive convolutional neural network," in WACV. IEEE, 2018, pp. 1113–1121.
[24] Lu, Xiankai, et al. "See more, know more: Unsupervised video object segmentation with co-attention siamese networks." Proceedings of the IEEE conference on computer vision and pattern recognition. 2019.
[25] R. R. Varior, B. Shuai, J. Tighe, and D. Modolo, "Scale-aware attention network for crowd counting," CVPR, 2019.

[26] N. Liu, Y. Long, C. Zou, Q. Niu, L. Pan, and H. Wu, "Adcrowdnet: An attention-injective deformable convolutional network for crowd understanding," CVPR, 2019.

[27] D. Onoro-Rubio and R. J. Lopez-Sastre, "Towards perspective-free ´ object counting with deep learning," in ECCV. Springer, 2016, pp. 615–629.

[28] L. Boominathan, S. S. Kruthiventi, and R. V. Babu, "Crowdnet: A deep convolutional network for dense crowd counting," in ACM MM. ACM, 2016, pp. 640–644.

[29] D. B. Sam and R. V. Babu, "Top-down feedback for crowd counting convolutional neural network," in AAAI, 2018.

[30] L. Liu, H. Wang, G. L. andWanli Ouyang, and L. Lin, "Crowd counting using deep recurrent spatial-aware network," in IJCAI, 2018, pp. 849– 855.

[31] M. Hossain, M. Hosseinzadeh, O. Chanda, and Y. Wang, "Crowd counting using scale-aware attention networks," in WACV. IEEE, 2019, pp. 1280–1288.

[32] A. Zhang, J. Shen, Z. Xiao, F. Zhu, X. Zhen, X. Cao, and L. Shao, "Relational attention network for crowd counting," in ICCV, 2019, pp. 6788–6797.

[33] Dadnet: Dilated-attentiondeformable convnet for crowd counting," in ACMMM. ACM, 2019, pp. 1823–1832.

[34] Sindagi, Vishwanath A., and Vishal M. Patel. "A survey of recent advances in cnn-based single image crowd counting and density estimation." Pattern Recognition Letters 107 (2018): 3-16.

[35] Shen, Zan, et al. "Crowd counting via adversarial cross-scale consistency pursuit." Proceedings of the IEEE conference on computer vision and pattern recognition. 2018.

[36] Shi, Zenglin, Pascal Mettes, and Cees GM Snoek. "Counting with focus for free." Proceedings of the IEEE International Conference on Computer Vision. 2019.

[37] Krizhevsky, Alex, Ilya Sutskever, and Geoffrey E. Hinton. "Imagenet classification with deep convolutional neural networks." Advances in neural information processing systems. 2012.